\documentclass[12pt]{article}
\RequirePackage[OT1]{fontenc}
\usepackage[english]{babel}
\usepackage{amsmath}
\usepackage{amssymb}
\usepackage{authblk}
\usepackage{graphicx,psfrag,epsf}
\usepackage{enumerate}
\usepackage{natbib}
\usepackage{threeparttable}
\usepackage{multirow}
\usepackage{stackengine}
\usepackage{setspace}
\usepackage{float}
\usepackage{appendix}
\usepackage{stackengine}
\usepackage{verbatim}
\usepackage[top=1in, bottom=1in, left=1in, right=1in]{geometry}
\usepackage{booktabs}
\RequirePackage[colorlinks,citecolor=blue,urlcolor=blue]{hyperref}

\usepackage{nameref}
\usepackage{varioref}
\usepackage{url}
\usepackage{comment}
\usepackage{lscape} 
\usepackage{cleveref}

\usepackage{color}
\definecolor{cmtcol}{rgb}{0.9, 0.36, 0}
\definecolor{cmtcol2}{rgb}{0.5, 0, 0.5}

\makeatletter
\newcommand\thankssymb[1]{\textsuperscript{\@fnsymbol{#1}}}
\makeatother

\setstackgap{L}{.7\normalbaselineskip}
\setstackEOL{\#}

\newcommand{\blind}{1}

\usepackage{xr}

\makeatletter
\newcommand*{\addFileDependency}[1]{
	\typeout{(#1)}
	\@addtofilelist{#1}
	\IfFileExists{#1}{}{\typeout{No file #1.}}
}
\makeatother

\newcommand*{\myexternaldocument}[1]{%
	\externaldocument{#1}%
	\addFileDependency{#1.tex}%
	\addFileDependency{#1.aux}%
}

\myexternaldocument{supplement}

\begin{document}

	\def\spacingset#1{\renewcommand{\baselinestretch}%
		{#1}\small\normalsize} \spacingset{1}

	
	\if1\blind
	{
		\title{\bf Leveraging text data for causal inference using electronic health records}
		\author[1]{Reagan Mozer\thanks{Corresponding author (email: rmozer@bentley.edu)}}
		\author[2]{Aaron R. Kaufman}
		\author[3]{Leo A. Celi}
		\author[4]{Luke Miratrix}

		\affil[1]{Bentley University}
		\affil[2]{New York University Abu Dhabi}
  		\affil[3]{Massachusetts Institute of Technology}
		\affil[4]{Harvard University}
		\date{}
		\maketitle
	} \fi
	
	\if0\blind
	{
		\bigskip
		\bigskip
		\bigskip
		\begin{center}
			{\LARGE\bf Using text data to improve causal analyses in medical research}
		\end{center}
		\medskip
	} \fi
	
	\bigskip
	\begin{abstract}

		In studies that rely on data from electronic health records (EHRs), unstructured text data such as clinical progress notes offer a rich source of information about patient characteristics and care that may be missing from structured data. Despite the prevalence of text in clinical research, these data are often ignored for the purposes of quantitative analysis due their complexity.
 This paper presents a unified framework for leveraging text data to support causal inference with electronic health data at multiple stages of analysis.
In particular, we consider how natural language processing and statistical text analysis can be combined with standard inferential techniques to address common challenges due to missing data, confounding bias, and treatment effect heterogeneity.
Through an application to a recent EHR study investigating the effects of a non-randomized medical intervention on patient outcomes, we show how incorporating text data in a traditional matching analysis can help strengthen the validity of an estimated treatment effect and identify patient subgroups that may benefit most from treatment.
We believe these methods have the potential to expand the scope of secondary analysis of clinical data to domains where structured EHR data is limited, such as in developing countries. To this end, we provide code and open-source replication materials to encourage adoption and broader exploration of these techniques in clinical research.
	\end{abstract}
	
	\noindent%
	{\it Keywords:}  text analysis, causal inference, observational studies, missing data
	\vfill
	
	\newpage
	\spacingset{1.5}

 \section{Introduction}
	\label{sec:intro}
	
	The field of medicine has been revolutionized in the last decade by the advent of large clinical databases \citep{jensen2012mining, friedman2013natural, evans2016electronic, cowie2017electronic}. Electronic health records (EHRs), administrative databases, and online registries hold a wealth of information with the potential to help answer long-standing questions across all facets of health care, from designing more effective treatment regimes \citep{komorowski2018artificial} to tailoring treatment to individuals based on their characteristics \citep{abul2019personalized}. 
	A primary challenge in this domain revolves around how to best harness that information, much of which is unstructured data such as text or complex data such as x-ray images \citep{koleck2019natural, huang2020fusion, tayefi2021challenges}. 

 In particular, the use of EHR data for causal inference presents a number of practical and methodological challenges. Chief among these is the threat of bias due to data quality and completeness issues. For example, discrepancies in how data is recorded across different EHR systems can lead to misclassification of treatment exposure and incomplete capture of outcomes \citep{joshua2022longitudinal}. A related issue is due to confounding by indication, where sicker patients (i.e., those most in need of treatment) will have more information captured in their records \citep{schneeweiss2005review}. This poses a significant challenge for causal identification in studies investigating the effectiveness of a medical intervention, as important confounders may only be observed for the treatment group. The heterogeneity of data types within EHRs, including structured and unstructured data, adds yet another layer of complexity to the data analysis pipeline.
	
	Despite these challenges, secondary analysis of even the lowest-hanging fruit from de-identified hospital records has, in recent years, yielded crucial insights into the effectiveness of medical interventions, both overturning received wisdom and improving patient care \citep{critical2016secondary}.
	These insights rest on a century of research in statistics, econometrics, and causal inference \citep{kleinberg2011review,hernan2019second} focusing on gleaning valid cause-and-effect relationships from observational data using analysis strategies such as interrupted time series, regression discontinuity, and matching.

	At the same time, a literature in the computational social sciences has opened new doors to quantify, analyze, and rigorously interpret unstructured data.
	Unsupervised techniques like Latent Dirichlet Allocation (LDA; \citealp{blei2003latent}) and the Structural Topic Model (STM; \citealp{roberts2014structural}) have enabled researchers with no \textit{a priori} hypotheses to easily measure the composition of text corpora, while supervised methods like Convolutional Neural Nets (CNN; \citealp{albawi2017understanding}) have allowed for the scalable and automatic production of document-level outcomes or covariates \citep{grimmer2021machine}, especially when traditional forms of quantitative data are lacking. 
	Recent work \citep[e.g.,][]{roberts2020adjusting, mozer2020matching, egami2022} has formalized notation and begun to introduce best practices for incorporating unstructured data such as text into observational studies for improving causal inference, showing how text can fruitfully improve causal estimates in domains such as internet censorship and media bias.

	In this paper, we consider how text data can be leveraged to tackle complications that commonly arise in EHR-based medical research due to (1) missing data, (2) non-randomized treatment assignment, and (3) heterogeneous treatment effects. Through a replication study, we demonstrate that text collected during routine medical practice can be used to enhance standard research workflows in each of these contexts.
	First, we show how text in the form of doctors' and nurses' notes, which are collected more frequently and contain far greater detail than typical patient-level quantitative data, can facilitate causal analysis in the presence of missing data. Second, we illustrate a method for incorporating text data into a matching analysis in order to strengthen the plausibility of causal claims in the context of non-randomized studies. Third, we show how to use text when estimating treatment effect heterogeneity: by conditioning on interpretable features of the text, we gain a more nuanced understanding of treatment efficacy and can identify subgroups that may be especially receptive to treatment (or especially harmed by it). And while our running example uses observational data, both our missing data imputation procedure and our heterogeneous treatment effects estimation method would apply in randomized experiments as well.

	This paper proceeds as follows.
	In Section~\ref{sec:motivating} we introduce our motivating example, an observational study investigating the efficacy of trans-thoracic echocardiography (a common diagnostic medical procedure), and outline our three applications of text data to improve causal analysis in this context. 
	We then present our data and associated notation, discussing how text data can be represented for statistical analysis, in Section~\ref{sec:data}.
	Next, we detail our methodologies for incorporating text into standard research workflows to improve missing data imputation (Section~\ref{sec:imputation}), improving the quality of a matching analysis (Section~\ref{sec:matching}), and identifying areas of impact heterogeneity  (Section~\ref{sec:hettx}). Within each of these sections, we outline our approach and present results from our case study.
	Section~\ref{sec:disc} closes with a discussion on how to think about using unstructured data such as text in empirical research more broadly, as well as links to our replication materials and open-source software.

	\section{Motivating application: the effects of transthoracic echocardiography on mortality among sepsis patients}\label{sec:motivating}

	In our motivating application, we seek to estimate the effects of transthoracic echocardiography (TTE), a non-randomized medical procedure used to create pictures of the heart, on patient-level outcomes using data collected from electronic health records.
	Like many procedures, TTE is both imprecise and expensive to perform.
	Furthermore, this diagnostic tool has been shown to change patient care strategies in as few as one-third of its patients \citep{matulevicius2013appropriate}.
	To contribute to ongoing research in this domain, we therefore seek to identify whether and for whom the use of TTE leads to meaningful improvement with respect to mortality and other clinical endpoints.
	
	As in all observational studies, we know that there may be systematic differences between the patients who receive TTE and those who do not (e.g., sicker patients may be more likely to receive the procedure) \citep{stuart2010matching}, so it is important to identify and control for those differences in order to precisely measure any treatment effect.
	A classic approach to conducting such an analysis, controlling for baseline differences, is to \emph{match} the data prior to analysis \citep{rosenbaum1983central}.
	In a matching analysis, such as that presented by \cite{feng2018tte}, researchers must first systematically identify treated and untreated subjects who are similar with respect to observed background characteristics, then remove patients without any matches.
	The final, matched, dataset is then arguably as-if randomized: we would expect any found differences in the outcome between these two groups to be due to the intervention in question (here, TTE) rather than inherent differences that existed before treatment because the treatment and the control group have similar values of those background characteristics.

	Unfortunately, the numerical data available to us to match with are imperfect -- several key variables have as much as 30\% missingness -- and so before we perform our matching step we may want to impute any observed missing values to avoid risking severe bias \citep{rubin1976inference, little2012prevention}.
	Best practice in imputation leverages the observed data to infer the missing values \citep{white2011multiple, azur2011multiple}; thus, incorporating richer covariate information may lead to more accurate imputations \citep{white2010bias}.
	For this, we turn to clinical text in the form of doctors' and nurses' notes \citep{assale2019revival, byrd2014automatic, rajkomar2018scalable}, which accompany the measured numerical data, to improve the imputation. These data provide a rich second view into patient well-being and can capture important aspects of the patients not represented by the numerical scores.
	
	The effectiveness of matching as a causal inference device rests on the claim that, once subjects who are similar with respect to the available, observed features have been identified, the treatment and matched control groups are ``as-if randomized.'' This is commonly called the \emph{selection on observables} assumption: we must assume that we are observing all the information necessary to ensure the treatment and control groups are truly comparable in terms of propensity to receive treatment \citep{rosenbaum2002observational}.
	Unfortunately, even with clean data, the selection on observables assumption is often weak: even thorough patient demographics, medical histories, and laboratory test results may not contain enough information to ensure that the treatment and control groups are as good as random. \citep{liu2013introduction}.
	To strengthen this assumption, we can again turn to the observed text. We argue that the doctors' and nurses' notes contain important information on the well-being of the patients not necessarily captured by the numerical data, and that by matching on the text in addition to the numerical data, we can bolster the assumption of no unmeasured confounders.
	
	Finally, after we use imputation to address the missingness in our data and conduct a text-reinforced matching analysis to calculate an overall treatment effect, we might be curious about \textit{heterogeneous} treatment effects \citep{fink2014testing, kent2018personalized, powers2018some, wendling2018comparing}: are there some subsets of patients for whom TTE is more effective than others? Are there subsets of patients for whom TTE is actively harmful?
	We can estimate treatment effects conditional on the observed numerical covariates, but as we have already established, these data may not be complete. To address this, we experiment with identifying heterogeneous treatment effects conditional on covariates derived \textit{from text}, and show that groups defined by text provide a richer and more powerful view of treatment effectiveness than the numerical covariates alone.

\begin{figure}[ht]

     \begin{center}
         \fbox{\includegraphics[width=1.0\textwidth]{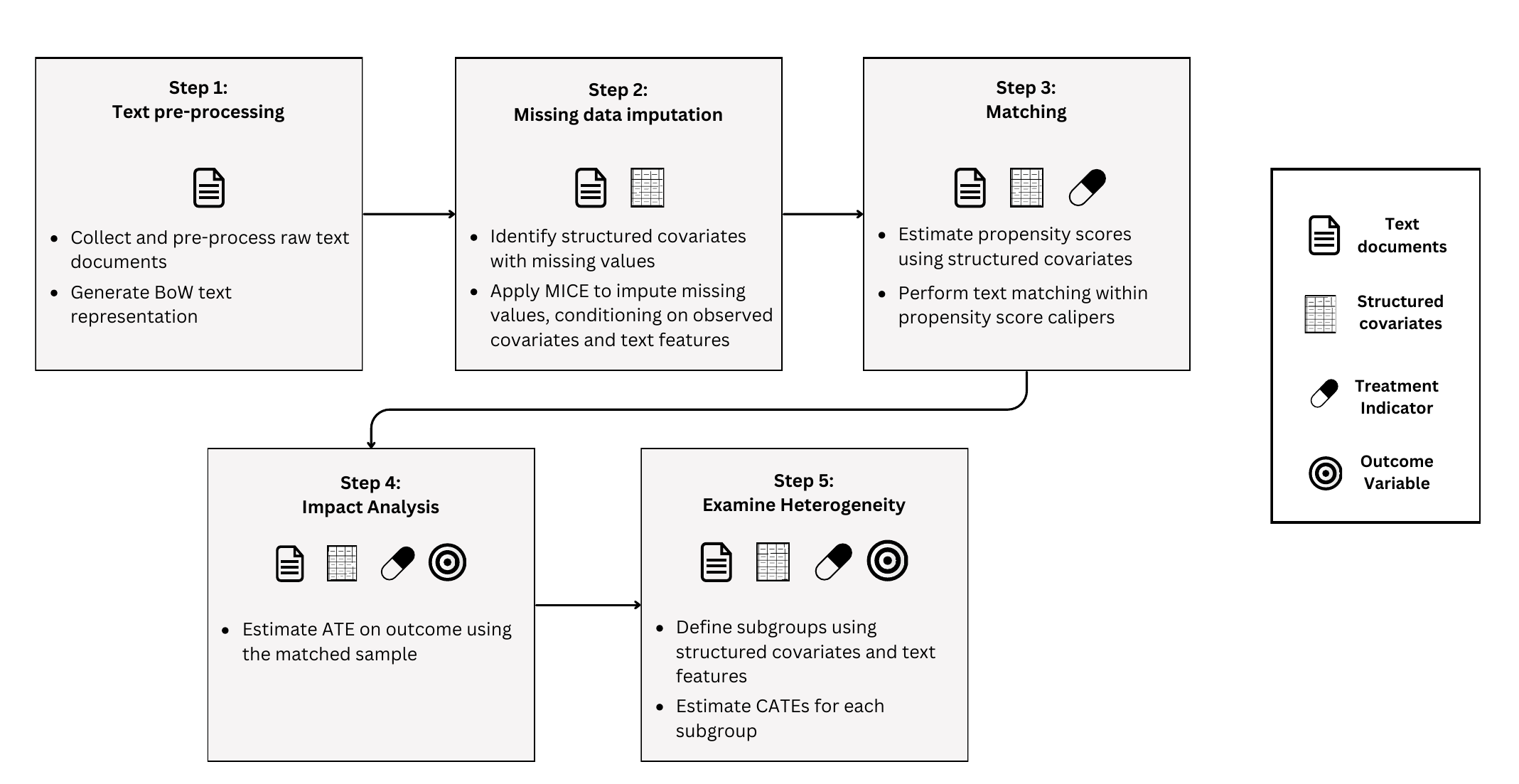}}
     \end{center}
     \caption{Analytical workflow for leveraging unstructured text data to support causal inference with EHR data.}
     \label{fig:flow_diag}
     \end{figure}

     Figure~\ref{fig:flow_diag} provides an overview of the analytical workflow we propose for leveraging unstructured text data to support this causal analysis. Our approach consists of five key steps: (1) text pre-processing to generate structured representations of the raw text data, (2) using text features to impute missing values in structured covariates, (3) incorporating text information into the matching procedure to strengthen the selection on observables assumption, (4) estimating the average treatment effect on the matched sample, and (5) examining treatment effect heterogeneity across subgroups defined by both the structured covariates and text features.

	At each step in the analysis pipeline, we argue that text data such as doctors' and nurses' notes can recover important, interpretable characteristics of the patients that are not captured within the numerical data, or are captured only with significant measurement error.
	We also argue that this problem is general: While large amounts of text data are often available in observational studies, these data are often unstructured, high-dimensional, and prone to inconsistencies and measurement errors (e.g., misspellings), and so are therefore typically ignored in causal analyses.
	Our key insight is that unstructured data such as text contain important information beyond what exists in tabular data \citep{wyss2023scalable}. We work to leverage such data, first showing that in our context, clinical text data do capture important information about patient health that can and should be used to make more precise inferences about the effects of medical procedures.

	\section{Study Design and Data}
 \label{sec:data}

 \subsection{Notation}
	Consider an observational study with $N$ units, indexed by $i=1,\ldots,N$.
	Let $Z_i$ be an indicator for treatment assignment which equals 1 for units in the treatment group and 0 for units in the control group.
	Following the principles and notation of the potential outcomes framework (Splawa-Neyman et al., 1923/\citeyear{SplawaNeyman:1990})
 , interest focuses on estimating differences between these groups on an outcome variable, $Y_i$, which takes value $Y_i(1)$ if unit $i$ is in the treatment group and $Y_i(0)$ if unit $i$ is in the control group. In particular, we are interested in estimating the average treatment effect (ATE) defined as $\tau=E[Y_i(1)-Y_i(0)]$. 
	
	For each unit we observe a vector of $p$ pre-treatment covariates $X_i = (X_{i1},X_{i2}, \ldots, X_{ip})$ as well as a text document $T_i$ containing a sequence of alphanumeric tokens (e.g., a word, a number, or punctuation mark).
	These documents are generally regarded as ``unstructured'' in the sense that their dimension is not well-defined.
	To address this issue, we impose structure on the text through a representation, $f(T)$, which maps each document to a finite, usually high-dimensional, quantitative space. Throughout this paper, we use the term \textit{structured covariates} to distinguish between the observed numerical variables $X_i$ and the set of numerical variables generated by applying a specific representation to the text. 

 \subsection{EHR Data Application}

	For our motivating application, we examine a subset 
 of the data first presented in \cite{feng2018tte}, which come from the Medical Information Mart for Intensive Care (MIMIC-III) database \citep{johnson2016mimic}. For a discussion of this data set and our inclusion criteria, see Appendix A of the Supplement.
	Our sample consists of 2,625 adult patients admitted with a diagnosis of sepsis in the medical and surgical intensive care units at a university hospital located in Boston, Massachusetts from 2001 to 2012. 
	Within this sample, the treatment group consists of 1,333 patients who received TTE during their ICU stay (defined by time stamps corresponding to times of admission and  discharge), and the control group is composed of 1,292 patients who did not receive a TTE during this time.  The primary outcome in this study is 28-day mortality from the time of ICU admission. 
	
	For each patient, we observe a vector of 45 structured pre-treatment covariates including demographic data, lab measurements, and other clinical variables. In addition to these structured numerical data, each patient is also associated with one or more text documents containing clinical progress notes written by physicians and nursing staff, as well as written evaluations from specialists at the time of ICU admission.
	Table~\ref{tab:descriptives} presents summaries of the patient characteristics in this sample as well as the number and type of documents observed for each patient.

\begin{table}
\caption{Characteristics of the ICU patients in the treatment and control groups. Values shown represent the mean (SD) for numerical variables and number of observations (percentage of sample) for discrete variables.}
\label{tab:descriptives}
\centering
\begin{tabular}[t]{lcc}
\toprule
 & Control & Treatment\\
 & $(N=1,292)$ & $(N=1,333)$ \\
\midrule
Age & 66.9 (17.4) & 65.79 (16.8)\\\addlinespace
Sex &  & \\
\hspace{5mm}Female & 644 (49.8\%) & 633 (47.5\%)\\
\hspace{5mm}Male & 648 (50.2\%) & 700 (52.5\%) \\\addlinespace
No. notes per patient & 2.62 (2.1) & 2.82 (2.3)\\
\hspace{5mm}Nursing notes & 0.67 (1.5) & 0.65 (1.6)\\
\hspace{5mm}Other notes& 1.3 (1.3) & 1.53 (1.4)\\
\hspace{5mm}Physician notes & 0.66 (1.3) & 0.63 (1.4)\\\addlinespace
Total text characters & 7182.44 & 7891.95 \\
 \hspace{5mm} per patient& (10569.3) & (12631) \\\addlinespace
Length of ICU stay & 4.85 (5.9) & 10.24 (10.4)\\\addlinespace
Ventilation free days & 18.83 (15) & 17.02 (17.4)\\\addlinespace
\bottomrule
\end{tabular}
\end{table}
	
	\subsection{Text pre-processing and representation}
	\label{subsec:rep}
 
    Raw text data requires preprocessing and transformation into a structured format to be suitable for quantitative analysis. 
    We considered two main approaches to representing the free-text clinical notes in our study (see Appendix B of the Supplement for a discussion of alternative text representations). 
    Our goal was to construct a set of numerical text-based covariates capturing information in the notes that may be relevant for (1) imputing missing values in the structured covariates, (2) identifying important potential confounders that should be adjusted for to allow for causal identification, and (3) uncovering heterogeneity in the estimated treatment effects.
    
    Our primary approach to representing the text is based on the classical bag-of-words (BoW) model \citep{salton1983introduction}, which views each document as an orderless collection of tokens (i.e., words, called unigrams, and multi-word phrases, called n-grams). 
    Under this model, each document is represented as a sparse vector of token counts $C_i=(c_{i1},\ldots,c_{id})$, where $c_{ij}$ denotes the number of occurrences of the $j$th token in document $i$ and $d$ denotes the size of the vocabulary. 
    These vectors are then collated into an $N \times d$ ``document-term matrix'' (DTM).
    
	Documents represented using the BoW model are usually pre-processed prior to analysis \citep{denny2018text}: extremely common words, extremely rare words, and punctuation are typically removed, and words sharing the same root are often combined (stemmed).
	For our application, we convert the text to lowercase and remove all punctuation and ``stop words'' (i.e., commonly used terms that contain little to no contextual information such as  ``the'' and ``and''). We then produce a DTM containing observed unigrams, bigrams, and trigrams.  
    
    In addition to the BoW model, we also explored an alternative text representation based on vector embeddings from a pre-trained large language model. Word embeddings are dense vector representations that capture similarities between words. In particular, we used the ClinicalBERT model \citep{huang2019clinicalbert}, which was trained on a similar corpus of medical texts collected from the MIMIC-III database. Using ClinicalBERT, each token in a corpus is represented as a 768-dimensional embedding vector, which we then aggregate to generate document-level summaries such that each document is represented as the occurrence-weighted average of the embeddings of all tokens contained within it. This produces, for each document in our corpus, a 768-long covariate vector.

    While the embedding-based representation has the potential to capture more complex semantic relationships between words and documents, for our application, we found that BoW representation to be better suited the task at hand (i.e., capturing clinically relevant terms and phrases in the text).

	\section{Imputing missing data with text}\label{sec:imputation}

	A common challenge facing researchers in the analysis of clinical data in both randomized and non-randomized studies is missing data \citep{rubin1991multiple, leyrat2019propensity, mainzer2022comparison}.
 In many studies, data may be completely missing for an important pre-treatment covariate or there may be missing values for one or more variables across different units.
 When left unaddressed, many inferential procedures will either exclude any units with missing covariate values (i.e., complete case analysis) or will drop any variables with missing values (i.e., complete variable analysis). 
 However, the validity of these approaches relies on the strong assumption that the data are missing completely at random (MCAR); that is, missingness is assumed to be independent of both the observed and unobserved data. Further, even when MCAR holds, 
failure to address this missingness can lead to a dramatic reduction in sample size, considerable information loss, or both.

	An alternative approach is to replace the missing values with a constant (e.g., the mean of the observed values for each covariate) and include missingness indicators as additional covariates. However, this can lead to biased results \citep{knol2010unpredictable}.
	Thus, the preferred method for addressing missing data in the context of causal analysis is multiple imputation, whereby each missing value is replaced by two or more plausible values \citep{rubin2004multiple}. This approach has been widely applied to handle missing data in medical studies \citep[e.g.,][]{barnard1999applications} and is known to have good statistical properties (i.e., yields unbiased and consistent estimators) under the more plausible missing at random (MAR) assumption. Under MAR, the observed data is sufficient to explain the missing data, and the missingness mechanism can be effectively ignored when estimating treatment effects.

	Multivariate imputation by chained equations (MICE, \citealp{buuren2010mice}) is a popular approach to multiple imputation that involves iteratively fitting a series of predictive models, whereby each partially-observed covariate is regressed on all other variables in the data. The missing values are then imputed from the estimated conditional distribution of each covariate. This process results in a set of completed (i.e. imputed) data sets that capture the uncertainty associated with predicting the missing values of each covariate. Matching can then be performed for each imputed dataset and the results combined to obtain an overall estimate of the treatment effect \citep{mitra2016comparison}. This approach has been shown to yield unbiased effect estimates when combined with propensity score analysis under the MAR assumption \citep{granger2019avoiding, leyrat2019propensity}.
 
	A key obstacle in implementing MICE is that the observed data may contain little information about the missing values.
	In this setting, text data may offer a useful source of supplemental information to enhance the imputation models and make the MAR assumption more plausible. 
Here, the dimensionality of text data is both a blessing and a curse. On the one hand, large amounts of text data may offer more valuable information about the missing variables, increasing our faith in MAR.
But as the dimension of the text grows, so too does the computational complexity of each conditional regression model.

 We propose using text as a way to \emph{augment} missing data imputation in settings where the data are assumed to be MAR.
 As with any missing data problem, researchers must still carefully consider the information available and conduct a rigorous assessment of the possible missing data mechanisms at play.
 We note that, while conditioning on observed text data may help strengthen the MAR assumption, it does not guarantee MAR will hold.
 Under our proposed approach, researchers must evaluate the plausibility of this assumption just as they would with a more traditional missing data enterprise.

	\subsection{Text imputation procedure}
	
	When we do not have complete numerical data $X$ for all units, but we do observe text $T$ for all units, we can use the observed text data to impute the missing covariate values needed for downstream analysis. 
	Following the literature \citep{d2000estimating, beesley2020mice}, 
 we advocate for applying multiple imputation without consideration of the outcome values so that any downstream analyses using the imputed values will not depend on assumptions about the outcome model. We note, however, that some studies have shown that including the outcome variable for imputation of missing covariates may be advantageous \citep{moons2006using,van2018flexible}.
	The goal is to use the observed text, together with the observed structured covariates, to infer the missing values for any partially observed numerical variables. By doing so, we can preserve a larger portion of our sample and leverage the full array of available information for downstream analyses.

    To incorporate text data into a multiple imputation effort, the first step is to generate a set of auxiliary variables (derived from summary measures of the text) that are believed to be predictive of the missing data.
    Once we have a set of text-based covariates, we can then condition on them in the MICE procedure just as with any other structured covariate. 
    While we could, in principle, use the full $d$-dimensional DTM of term counts as our set of text-based covariates, this may be inefficient and computationally burdensome in settings where the observed text spans a large vocabulary.
    Instead, we propose using multinomial inverse regression (MNIR), a dimension-reduction technique introduced by \cite{taddy2013multinomial}, to generate a low-dimensional representation of the text that incorporates information from the observed numerical covariates. 
    MNIR is an extension of multinomial logistic regression that aims to find a low-dimensional representation of the DTM that best explains the variation in the structured covariates.

 Given a DTM constructed under the bag-of-words model (denoted by $C$), MNIR performs a series of $d$ separate regularized regressions, one for each term in the vocabulary, by regressing the columns of $C$ onto the observed structured covariates $X$. This process results in a $d\times p$ matrix of estimated regression coefficients, denoted by $\Phi$, which are then used to generate a $p$-vector of ``projections" for each text document through the linear mapping $S_i=C_i\Phi/m_i$, where $m_i=\sum_j c_{ij}$ is the total word count for document $i$. These projections are referred to as ``sufficient reduction'' (SR) scores and are sufficient statistics in the sense that the original covariates $X_i$ are independent of the text counts $C_i$ given the SR projections $S_i$ \citep{taddy2013multinomial, taddy2015distributed}. Each column of $S$ therefore provides a univariate summary of the observed text that reflects the variance of the corresponding structured covariate.
	Using the above elements, our text-based imputation procedure is as follows.
	First, using only the complete cases within our data, we apply MNIR to the full DTM to estimate a set of regression coefficients $\hat{\Phi}$ that capture the relationship between the observed text and each structured covariate (assuming a MAR missingness mechanism).
	Next, we calculate the $p$-vector of SR projections $\hat{S}_i=C_i\hat{\Phi}/m_i$ for each document in the full sample (including observations with one or more missing values for the structured covariates). 
	Third, we bind the full data set of structured covariates, $X$, to the matrix of SR scores for these covariates, $\hat{S}$.
	Now, for each initial structural covariate in our data set, we have the observed values, which may have some missingness, and a corresponding text-based projection, which is fully observed.
	
	Finally, we apply MICE, implemented using the \texttt{mice} package in \text{R}, to impute the missing values of the structured covariates using the augmented covariate matrix $X^{\star}=(X, \hat{S})$.
	Information from the text, insofar as we can discover how it relates to the structured covariates from the observed data, can then be used to give, in effect, a proxy value for each covariate that can be exploited by the MICE procedure.

	\subsection{Results: Text improves imputation quality}\label{sec:imputation_results}
	
	Our data contains numerical measurements for 45 structured covariates including demographic information and clinical and laboratory measurements for each patient. Among these, 15 variables contain missing values. The mean and median missingness proportions across variables are 10\% and 3\%. 
	
	These 15 partially-observed variables have been excluded in previous analyses of the data \citep{feng2018tte}. Standard practice is either to consider only complete cases (listwise deletion) or to use MICE to impute the missing values for each pre-treatment covariates using a model that conditions on the observed covariates.

	To compare the performance of imputation techniques that include text-based covariates compared to those that ignore the text data, we conduct a cross-validation procedure in which we predict held-out values of each covariate and attempt to impute them using both standard MICE and our MICE procedure augmented with text data.
	
\begin{figure}[ht]
		\begin{center}
			\includegraphics[width=0.75\textwidth]{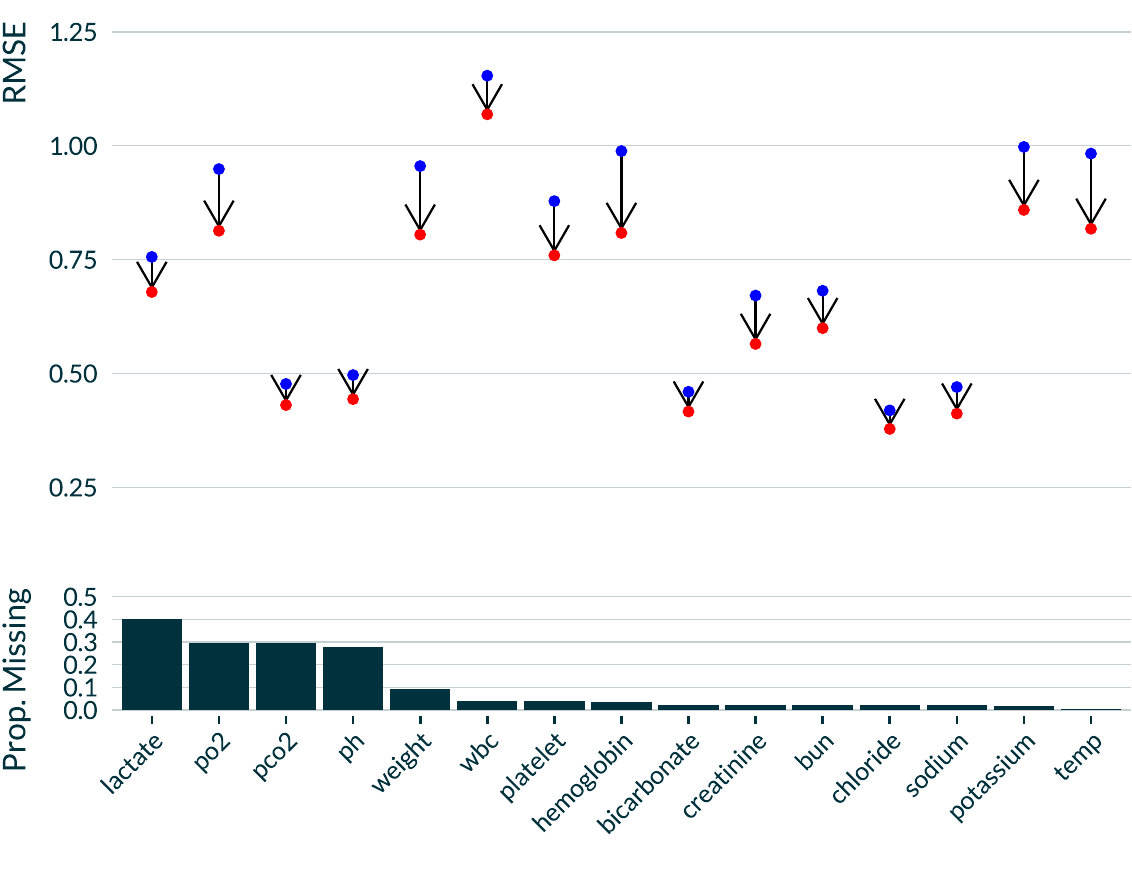}
		\end{center}
		\caption{Performance of the multiple imputation models for each covariate when conditioning on both the observed structured covariates and text-based covariates (red) compared to the structured covariates alone (blue).}
		\label{fig:missing}
	\end{figure}

	For each missing variable, we fit a regularized linear regression under two different model specifications. In the first model, we regress the target variable on all other numerical covariates (standardized prior to model fitting) and ignore the text data. In the second model, we regress the target variable on all numerical covariates plus the set of text projections estimated using MNIR. For each variable, both models are estimated using 10-fold cross-validation (CV). We then compare the performance of the two models based on the root mean squared error (RMSE) and squared correlation ($R^2$) achieved by each model across all testing samples.
	Figure~\ref{fig:missing} and Table~\ref{tab:missing} show the results from this procedure.

\begin{table}
\caption{Performance of regularized regression models trained to predict covariates with missing values as a function of only structured vs. structured and text-based covariates.}
\label{tab:missing}
\centering
\begin{tabular}{lcccclc}
\toprule
\multicolumn{1}{c}{} & \multicolumn{1}{c}{} & \multicolumn{2}{c}{$R^2$} & \multicolumn{3}{c}{RMSE} \\
\cmidrule(l{3pt}r{3pt}){3-4} \cmidrule(l{3pt}r{3pt}){5-7}
Variable & \# Missing & \Longstack{Structured \#Only} & \Longstack{Structured\# + Text} & \Longstack{Structured \#Only} & \Longstack{Structured\# + Text}& \Longstack{Diff. \#(\% change)}\\
\midrule
Creatinine & 61 & 0.55 & 0.69 & 0.67 & 0.55 & 0.12 (17.6\%)\\
Weight & 250 & 0.18 & 0.45 & 0.96 & 0.80 & 0.16 (16.9\%)\\
Temp & 10 & 0.23 & 0.47 & 0.98 & 0.81 & 0.16 (16.8\%)\\
Hemoglobin & 95 & 0.13 & 0.39 & 0.96 & 0.81 & 0.16 (16.4\%)\\
Potassium & 47 & 0.20 & 0.42 & 0.99 & 0.84 & 0.15 (15.0\%)\\
PO2 & 777 & 0.16 & 0.39 & 0.95 & 0.81 & 0.14 (14.9\%)\\
Sodium & 54 & 0.78 & 0.82 & 0.47 & 0.41 & 0.06 (12.3\%)\\
Platelet & 101 & 0.21 & 0.39 & 0.87 & 0.77 & 0.11 (12.1\%)\\
BUN & 60 & 0.56 & 0.65 & 0.68 & 0.60 & 0.08 (12.0\%)\\
Lactate & 1057 & 0.47 & 0.57 & 0.76 & 0.67 & 0.08 (10.8\%)\\
PH & 731 & 0.76 & 0.81 & 0.49 & 0.44 & 0.05 (10.7\%)\\
Bicarbonate & 64 & 0.81 & 0.84 & 0.46 & 0.42 & 0.05 (10.0\%)\\
PCO2 & 777 & 0.77 & 0.81 & 0.47 & 0.43 & 0.05 (10.0\%)\\
Chloride & 57 & 0.82 & 0.85 & 0.42 & 0.38 & 0.04 (9.1\%)\\
WBC & 102 & 0.10 & 0.24 & 1.07 & 1.01 & 0.06 (5.7\%)\\
\bottomrule
\end{tabular}
\end{table}

	Across all 15 variables with missing values, we find that including text in the imputation procedure meaningfully improves predictive accuracy: reduction in RMSE ranges from 5.7\% to 17.6\%. $R^2$ improves from 0.03 (3.7\%) to 0.28 (154\%). These improvements are only weakly related to the amount of missingness, indicating that this procedure is useful even when missingness is high. The number of missing values correlates with the reduction in RMSE at -0.23 and with the increase in $R^2$ at -0.15. Overall, these results suggest that conditioning on the observed text can enhance the predictive performance of our missing data models, allowing for more accurate imputations that can then be used for downstream analyses.

	\section{Matching with text data}\label{sec:matching}

	A typical matching study using electronic health records seeks to estimate the observational effect of a treatment, perhaps a procedure or a drug, on a clinical outcome like whether the patient recovers, is transferred out of Intensive Care within 10 days, or whether the patient survives at all.
	A critical issue in these contexts is that patients who are most in need of treatment are the most likely to receive it. 
	These patients are also the most likely to incur a negative outcome regardless of treatment.
	In this scenario, failure to control for patients' baseline health and pre-treatment characteristics can introduce bias in the estimated treatment impact.
	
	In our motivating example, for instance, if patients in critical condition are more likely to receive treatment with TTE, their higher mortality rates may be attributed to the treatment itself rather than their initial health status. Conversely, patients whose condition rapidly improves may be discharged before treatment with TTE has been considered, leading to an underestimation of the treatment impact.
	Therefore, it is crucial that we identify and appropriately adjust for any potentially confounding factors such as health severity at the time of admission in order to obtain an unbiased estimate of the true treatment effect.
	
	To correct for these sources of bias, researchers might match on patients' pre-treatment conditions.
	Patients who are very healthy might never receive the treatment, while those who are very unhealthy might \textit{always} receive it; these observations are removed from the data set since there are no similar patients with the opposite treatment condition.
	The remaining observations, once matched like with like, can then approximate a randomized experiment. 
	
	Dozens of matching procedures exist to create treatment and control groups that are more similar while minimizing the proportion of observations that get thrown out \citep{stuart2010matching}.
	However, matching methods, especially model-dependent methods such as propensity score matching, are only as strong as the covariates available to match on \citep{austin2008critical, caliendo2008some}.
	If there are important variables confounding the relationship between treatment and outcome that are not observed in the data set, researchers will be unable to estimate an unbiased treatment effect.
	This is precisely how we use text to improve causal inference: text data may contain information about important confounders that may not appear in structured data, and incorporating this text into matching can reduce the bias caused by omitted confounders. See Appendix C of the Supplement for a discussion of the identifying assumptions required for matching with text data.

	\subsection{Text matching procedure}
	In recent years, there has been a growing body of work that considers how matching methods can be used to adjust for high-dimensional text-based confounders when conducting an impact analysis using observational data \citep[e.g.,][]{roberts2020adjusting,keith2020text}. 
For instance, a recent study by \cite{mozer2020matching} used a subset of data from the same observational study by \cite{feng2018transthoracic} to examine the effects of TTE on survival outcomes. In that study, the authors adjusted for confounding using text collected from nurses intake notes. Here, we extend the analysis by \cite{mozer2020matching} by examining a larger patient sample and incorporating additional sources of text, including clinical notes written by physicians, medical specialists, and other hospital staff at the time of ICU admission.
	
	To understand the benefit of using text to improve match quality in this setting, we apply two different matching procedures to this data: one that matches patients only on numerical covariates and ignores the text data, and one that matches patients using both the numerical and text-based covariates.
	In the first procedure, we match treated and control units using optimal full matching on estimated propensity scores, which are calculated by fitting a logistic regression of the indicator for treatment assignment (receipt of TTE) on the observed numerical covariates.
	We enforce a propensity score caliper equal to 0.1 standard deviations of the estimated distribution, which discards any treated units for whom there are no control units within a suitable distance.
	
	In the second procedure, we perform text matching within propensity score calipers defined by the numerical propensity score.
	That is, for each treated unit, we first identify a set of possible controls defined by closeness on the numerical propensity score.
	We then conduct optimal full matching using the cosine distance calculated over the DTM to find the set of matched samples with the smallest average absolute distance across all matched pairs.
	
	Prior to matching, we leverage the text-based imputation technique described in Section~\ref{sec:imputation} to multiply impute the missing values for each partially observed quantitative covariate. Following the ``within" approach proposed by \citet{mitra2016comparison}, we first generate five completed versions of the dataset, each containing no missing values across the quantitative covariates. We then implement the matching procedure within each of these completed datasets and aggregate the corresponding covariate balance statistics and treatment effect estimates using standard combining rules \citep{rubin2004multiple}. This approach has been shown to be superior to approaches that aggregate across multiply imputed data sets prior to matching \citep{leyrat2019propensity, penning2016comments, granger2019avoiding}.

    We note that, although one might initially be concerned about using the same text in multiple stages (sometimes called the ``double-dipping'' problem \citep{ball2020double}), we believe this to not be a concern in our context. As long as imputation is restricted to baseline covariates, the resulting imputed dataset will also be ``baseline.''
	Similarly, as long as matching does not use outcomes to form matches, researchers are free to match however they want, and can even refit propensity score models after looking at covariate balance post-matching and add or remove variables to improve measures of match quality, and so forth.

	\subsection{Results: Text improves covariate balance}\label{sec:matching_results}

	In our motivating application, we first examine the baseline covariate balance between the treatment and control groups for the full (unmatched) sample to serve as a basis for evaluating our matching methods. As a measure of balance, we computed the standardized difference in means between the treatment and control groups with respect to each of the 45 structured covariates. We also evaluated balance for an additional structured variable indicating the presence of missing data, which was equal to 1 for patients with any missing values in the structured covariates and 0 for patients with no missing values. To evaluate baseline balance with respect to the text, we also constructed a set of 30 text-based covariates measuring the prevalence of a set of key terms and phrases capturing clinically-relevant information, as determined by clinicians.
	These covariates represent important potential confounders that are not included in the structured data but are likely to be captured within the clinical text notes.

\begin{figure}[ht]
		\begin{center}
			\includegraphics[width=0.85\textwidth]{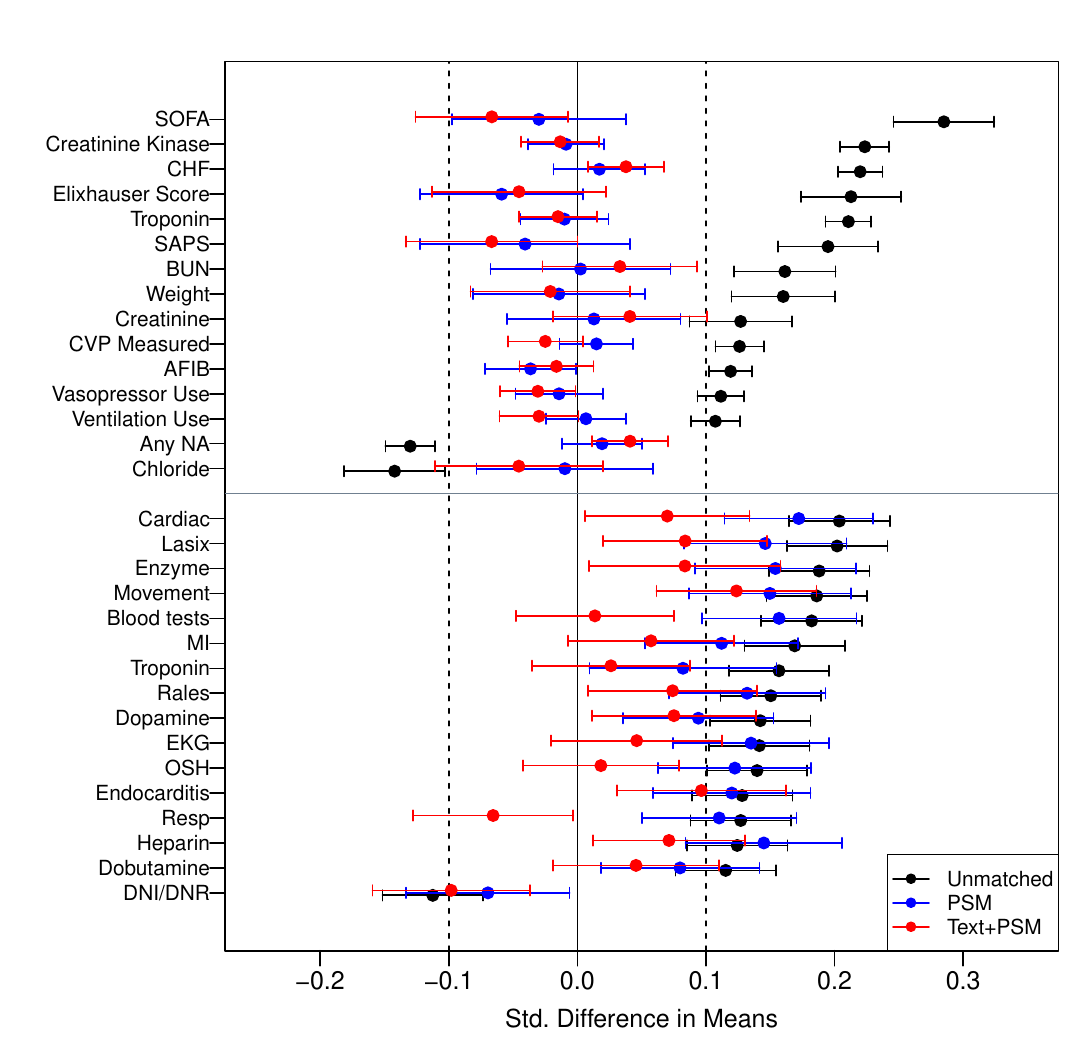}
		\end{center}
		\caption{Standardized differences in means between treatment and control groups for 15 structured baseline covariates (top) and 16 text-based covariates (bottom) before matching (black), after propensity score matching (red), and after text matching (blue). 
  Point estimates and 95\% confidence intervals are aggregated across 5 multiply imputed versions of the structured covariates. Text matching maintains original balance on structural covariates and improves balance on text covariates.}
		\label{fig:mimic1}
	\end{figure}

    Examining the baseline covariate balance for the full unmatched sample, we found notable imbalances (i.e., a standardized difference in means greater than $\pm$ 0.1) between treatment groups for 15 of the 46 structured covariates and 16 of the 30 text-based covariates. 
    Figure~\ref{fig:mimic1} shows the standardized differences in means between the treatment and control groups for the set of of structured and text-based covariates with large baseline imbalances before matching, after propensity score matching (PSM) on numeric covariates alone, and after cosine matching on the DTM within propensity score calipers.
	Table~\ref{tab:matching} summarizes the survival rates in the treatment and control groups within each sample.

\begin{table}[htbp]
\label{tab:matching}
  \centering
\caption{Survival rates for treatment and control groups and estimated treatment effects before and after propensity score matching (PSM) and text matching within propensity score calipers.}
\begin{tabular}{lcccl}
\hline
    \multicolumn{1}{c}{\textbf{Method}} & \multicolumn{1}{c}{\textbf{Effective}} & \multicolumn{2}{c}{\textbf{Survival Rate}} & \multicolumn{1}{c}{\textbf{Difference}} \\\cline{3-4}
          & \multicolumn{1}{c}{\textbf{Sample Size}} & \multicolumn{1}{c}{\textbf{Control}} & \multicolumn{1}{c}{\textbf{Treatment}} & \textbf{(Std. Error)} \\
          \hline
    Before matching & 1292  & 72.1\% & 72.7\% & 0.6\% (2.6\%) \\
    PSM   & 945   & 63.4\% & 72.6\% & 9.2\% (2.8\%) \\
    Text matching & 1001  & 65.1\% & 72.6\% & 7.5\% (2.7\%) \\
    \hline
    \end{tabular}%
\end{table}%
	
	These results highlight the importance of conditioning on \textit{all} available data when making inferences using observational data. In particular, while PSM is able to adequately balance the structured covariates, it fails to sufficiently adjust for differences between treatment and control groups on a number of important potential confounders within the text. For instance, the unmatched sample shows large imbalances between treatment and control groups in their use of the term ``cardiac" as well as their total number of references to common blood pressure medications including Lasix, nicardipine, and dobutamine. Matching on the estimated propensity scores reduces these imbalances only slightly, while text matching within propensity score calipers considerably improves balance with respect to these variables. Incorporating the text data into the matching procedure leads to similar improvements in balance for the other text-based variables, while also maintaining a suitable overall balance on the numerical covariates. Notably, text matching leads to a treatment effect estimate that is similar in magnitude to that produced by PSM, but with a slightly smaller standard error. These results support previous findings suggesting that the use of TTE in critical care settings leads to significant improvement in survival outcomes \citep{feng2018tte}. The reduction in standard error observed for the text-matched sample also showcases how text can be used to enhance the precision of impact analysis in this setting.
	
	By incorporating text data into our matching analysis, we are effectively strengthening the assumption of selection on observables, which implies that there is no unobserved confounder that could explain away the estimated treatment effect.  In our case, conditioning on summary features extracted from clinicians' intake and progress notes allows us to capture more information about patients' baseline health status and initial prognosis that are believed to affect both the treatment assignment and our outcome of 28-day mortality. While it is not possible to test this assumption directly, one could potentially conduct a sensitivity analysis to gauge how the inclusion of text in the matching step affects the sensitivity of our results to unmeasured confounding. To conduct a sensitivity analysis, one possibility would be to follow the approach of \cite{rosenbaum2002observational}, which involves specifying a model for the potential outcomes and the assignment mechanism as well as a sensitivity parameter $\Gamma$ that represents the degree of potential hidden bias in our data. In our context, we would need to specify two separate models: one that uses only the structured covariates and another that uses both the structured and unstructured (text) data. For each model, we could then evaluate the robustness of our findings by systematically varying $\Gamma$ and assessing how the magnitude of our estimated treatment effect changes. Finally, the difference in sensitivity parameters between the two models could provide a measure of the extent to which text data reduces the sensitivity of our results to unmeasured confounding.

	\section{Text-based treatment effect heterogeneity}\label{sec:hettx}

	Treatment effect heterogeneity is at the core of ``precision medicine'' \citep{koenig2017precision}, where medical procedures are targeted to those most likely to benefit, as predicted by their traits.
	The decision rules behind precision medicine are typically estimated with large historical databases of patients' traits, treatments, and outcomes.
	For example, some precision medicine seeks to identify genetic markers \citep{ashley2016towards} or even socio-behavioral patient-level predictors \citep{riley2015news} that predict likely response to treatment.
	By identifying \textit{ex-ante} which types of patients are more likely to benefit from specific treatments, clinicians can reduce the costs, and health risks, of ineffective procedures.
	For example, researchers might find through mining electronic health records that a procedure works best on younger patients, female patients, or, at the extreme, young female patients within a certain bandwidth of serum creatinine as measured by a lab test.

	In our case, considering our matched data as an as-if randomized experiment, we look for aspects of the text data that predict which patients are more or less likely to respond to treatment.
	Especially in data-poor environments where patient demographics and lab data are sparse, doctors' and nurses' notes may contain important information that could help us identify subgroups of patients especially amenable to treatment.
	These notes often contain details that would be difficult to capture in tabular format: measures of patient health, activity level, and mood, or facets of the physician's knowledge and skill, all subjective variables that may predict greater sensitivity to treatment.

	\subsection{Estimation procedure}
	
	Identifying the \textit{best} subgroups to administer treatment is challenging: we can only identify conditional treatment effects for covariates we measure, and it is impossible to know \textit{a priori} which covariates might be important without strong theoretical guidance.
	We thus are faced with a model-selection or multiple-testing problem: we explore a range of candidate ways of grouping patients and try to identify those ways that best separate the more- and less-treatment-responsive.

 	We consider five approaches to locating heterogeneous treatment effects: one baseline method that does not use text and four methods that use text in different ways.
	
	Our baseline approach groups patients according to values of the structured covariates. For each of these covariates, we first fit a receiver operating characteristic (ROC) curve to the relationship between that variable and our outcome of 28-day mortality. We then identify the value of that covariate that optimizes the ROC curve based on Youden's index (J) \citep{youden1950index} and use that value as a discretizing threshold, partitioning the covariate into two non-overlapping subgroups.

 For each of partition of our data into two subgroups, we estimate the difference in average impact between the two subgroups as the interaction effect between the treatment and subgroup indicators.
 In particular, for each partition in turn, we fit a linear regression with an interaction between the treatment indicator and this subgroup indicator, using unit-level weights derived from the full matching procedure (to maintain the quasi-experimental balance of our data).
 We calculate standard errors considering the matched groups as blocks, following \citep{pashley2021insights}; these robust standard errors also allow for any heteroskedasticity.
 
 Once done, we have a point estimate, standard error, and $p$-value for each interaction effect for each partition.
 This is a massive multiple-testing problem for the 45 baseline covariates, and even more so for the 5,653 text-based covariates described below.
 To reduce spurious findings, we use a false discovery rate correction on the set of estimated interaction effects to locate significant findings and adjust all point estimates using a shrinkage estimator; see Appendix D of the Supplement for additional details.

 Our text-based methods share the same spirit as the baseline method: we use our text data to generate subgroups of interest and then assess whether those subgroups have different average impacts.
 Each text-based method offers an alternate way of making subgroups out of the text.
 Given the richness of the text, regardless of which approach we take to making covariates, we end up with many possible groupings of patients, so we again use a multiple testing correction to control for that broad exploration.
 
Our first text-based method uses the bag-of-words representation for each patient, treating the presence of each text token (unigrams, bigrams, and trigrams) in the notes for a patient as a separate covariate, and searches across these text-based covariates for treatment effect heterogeneity.
In particular, for each text token, we create an indicator variable for the presence of that token in a patient's clinical notes, and then fit a linear regression with an interaction between the treatment indicator and this subgroup indicator, as described above for the base covariates.

The bag-of-words method is pleasantly interpretable: we might find, for example, that patients whose notes mention the word ``lethargic'' have stronger treatment effects than those whose notes do not, suggesting that lethargic patients are a good target population for the treatment. In short, this method uses nearly raw text data to effectively provide additional important patient covariates that should be included in the structured data but are not.
	
	The remaining three text methods aggregate the text before looking for treatment variation, which is more holistic in terms of the text but is less interpretable.
	Method two uses PCA to perform dimension reduction on the document-term matrix, discretizes the loadings of the principal components using an optimal threshold, and then tests for interactions between the level of each principal component and treatment assignment.
	This method automatically identifies linear combinations of words that might better identify clusters of patients, though at the cost of interpretability.
	
	The third and fourth methods use a large language model called ClinicalBERT \citep{alsentzer2019publicly, huang2019clinicalbert} that has been fine-tuned on clinical notes. We use these embeddings to represent the doctors' and nurses' notes in a relatively low-dimensional space. In method three we discretize the elements of this embedding vector using the same approach as with our structured covariates, identifying the value of that element that optimizes the ROC curve between the variable and the outcome and partitioning the variable according to that threshold.
 Our fourth approach instead performs additional dimension reduction on the embedding vectors using PCA and identifies the optimal discretizing threshold among the loadings of the principal components.
 
	Ultimately, we find the first text-based method, representing doctors' and nurses' notes as a bag-of-words vector and then, for each $n$-gram, comparing impacts for patients with or without that token in their charts, is superior to the other text-based representations we explored (see Appendix D of the Supplement for further discussion).
	The bag-of-words method has three key advantages: (1) it is conceptually straightforward, (2) it is computationally easy, and (3) it is interpretable.
	By avoiding reliance on large language models like ClinicalBERT, our preferred method avoids long computation time and black box models even though it estimates an interacted treatment impact for each text token in the corpus
	By describing treatment variation in terms of differences in the use of plain-text words or $n$-grams, our preferred method also allows us to more easily interrogate the results we find to ensure that the heterogeneous effects are conditioned upon substantively meaningful dimensions.

	\subsection{Results: Text reveals important heterogeneity}\label{sec:hettx_results}
	
\begin{figure}[!h]
		\begin{center}
			\includegraphics[width=0.75\textwidth]{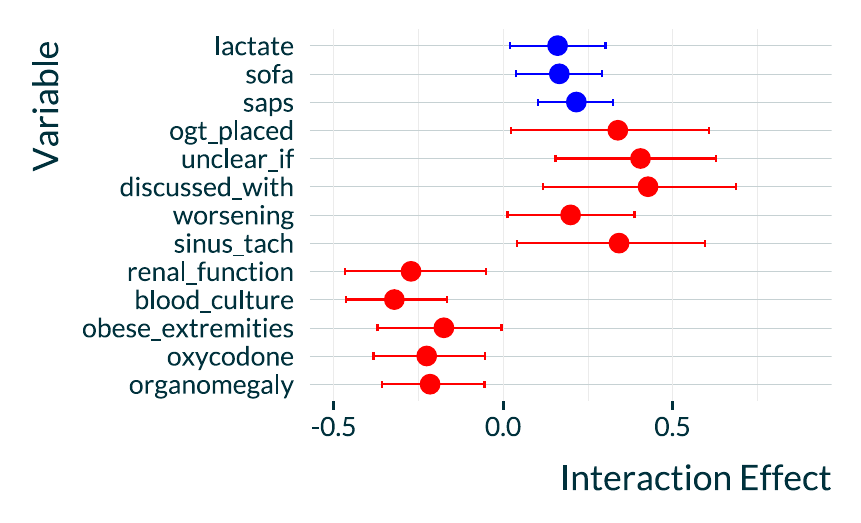}
		\end{center}
		\caption{Interaction Effects for thirteen variables. Variables in blue are determined by optimal cutoffs of structured covariates; variables in red are determined by the presence or absence of given text features. Text features identify groupings with comparably large or larger differences in impact than structured covariates.}
		\label{fig:hettx}
	\end{figure}

	Figure~\ref{fig:hettx} shows the final estimated interaction effect for thirteen covariates that survive our false discovery screen and shrinkage procedure.
 We find notable heterogeneity in the estimated treatment effects across these subgroups. The first three, in blue, are the structured covariates with the strongest heterogeneous treatment effects. The next ten, in red, are text-based covariates: five with the strongest \textit{positive} heterogeneous treatment effects, and five with the strongest \textit{negative} effects. In total, 75 variables, 3 structured and 72 text, have statistically significant interaction effects after shrinkage. 
	
	For example, the second and third variables are SOFA and SAPS, two common mortality prediction scores based on patients' clinical status. We see that patients with high SOFA and SAPS scores have interaction effects of $+0.166$ and $+0.216$, indicating that those subpopulations are good candidates for TTE. 
	
	The four of the next five text-based variables, however, all have much stronger heterogeneous effects than the structured covariates. For patients whose notes have text token ``sinus tach,'' TTE has a conditional effect of 0.344, more than ten times larger than the overall effect and more than twice as large as the interaction term on SOFA. The tokens ``unclear if'', ``discussed with'', ``worsening'', and ``outside hospital'' all have similarly large effects. Intuitively, these five terms all indicate patients in severe condition, especially related to the heart: ``ogt placed'' refers to mechanical ventilation, a common remediation for heart failure, and ``sinus tach'' indicates dangerously elevated heart rate, while  ``discussed with'' results from consulting outside physicians. Overall, there are fifteen text-based variables with stronger interaction effects than the strongest structured covariate.
	The final five text-based variables have \textit{negative} conditional effects: patients whose notes contain the terms ``renal function'', ``obese extremities'', ``solumedrol'', ``oxycodone'', or ``organomegaly'' appear to have \textit{worse} outcomes when receiving TTE compared to patients who did not receive TTE, with negative effects approximately ten times larger than the overall positive effect. These terms all indicate severe conditions that are \textit{not} related to the heart, suggesting that for these patients, TTE took vital time their doctors could have used to provide more appropriate care. There are fourteen variables with stronger negative effects than the strongest structured variable's positive effects.
	
	Importantly, while terms like ``obese extremities'' and ``renal function'' may be highly correlated with lab tests like weight, blood pressure, and serum creatinine, their presence in doctors' and nurses' notes is important information in itself. These notes are short and doctors' and nurses' time is valuable, so only patients for whom those lab results are notable will appear in the text data.

	\section{Discussion}\label{sec:disc}
	
	Text data, routinely collected and often ignored, can supplement and improve data analysis and causal inference in three separate phases of research: it can bolster a multiple imputation procedure to reduce error in resolving missing data problems, it can reduce the space of unobserved confounding variables when matching to identify causal effects, and it can act as subgroup covariates to identify individuals most responsive to treatment.
	Usefully, text can be used for all three purposes within the same research project.
	We note, however, all text data used within this framework should collected prior to treatment. Otherwise, using this text for imputation and/or matching will produce biased results \citep{blackwell2018make}. In some cases, it may be difficult to know when the text was produced relative to treatment; in other cases, text may be mislabeled as produced pre-treatment when it was not. For these reasons, it is critical to engage with data producers and understand what types of errors are likely to occur.

     In our empirical example investigating the effects of TTE, a non-randomized medical intervention, on patient outcomes, applying the proposed methods yielded several important insights. 
     First, leveraging information from clinical notes in the missing data imputation procedure improved the accuracy of our imputation models, strengthening the validity of downstream inferences based on the imputed values. 
     Second, by incorporating text features into the matching procedure, we were able to construct more balanced comparison groups and increase the precision of our estimated treatment effect. 
     Finally, conditioning on features extracted from the text revealed considerable heterogeneity in the effect of TTE across various patient subgroups. 
     In particular, our results suggest that TTE was most effective for patients whose clinical notes contained mentions of acute conditions like heart failure. 
     While more work is needed to validate these findings, these findings highlight the potential of text data to provide a more nuanced understanding of treatment effectiveness and effect heterogeneity.

    An important limitation of our application to the study of TTE is the lack of an active comparator. In particular, for studies that rely on secondary health data (e.g., from electronic medical records, administrative databases, or health registries),  the potential for confounding by indication poses a major threat to causal inference \citep{poses1995controlling}. This issue arises when treatment is given primarily to the sickest patients (i.e., those with clear indications for treatment), who are also the most likely to have adverse outcomes. As a result, there may be systematic differences in the type and amount of data captured for treated versus untreated patients, leading to biased and/or misleading effect estimates \citep{joshua2022longitudinal}. 
    The active comparator design attempts to address this issue by comparing patients who received the treatment of interest to a set of controls who received an alternative treatment based on similar indications \citep{lund2015active, yoshida2015active}. 
    In our case, no such active comparator was available. Thus, while we took steps to ensure consistent availability of information across groups (e.g., we used only pre-treatment text data from the date of ICU admission), our effect estimates may still be biased. However, the primary goal of this paper is to present a range of techniques for using unstructured text data to support causal analyses, with the TTE study serving as an illustrative example.

	More broadly, we believe our proposed methods can help expand the use of text data in observational studies and beyond, for example in both the design and analysis of randomized experiments. For instance, we might establish inclusion and exclusion criteria for enrolling units into an experiment based on a qualitative (i.e. textual) assessment of their symptoms and psychological characteristics. Alternatively, we may use text data for covariate adjustments in the analysis phase of an experiment by estimating treatment effects that condition on features of the text.

	We are optimistic that these advances can help distribute the fruits of the medical data revolution to the developing world. It is clear that many results derived from hospitals in the United States do not generalize well to other countries or different populations; moreover, data limitations preclude replicating the same studies in those broader contexts. By innovating methods that leverage the data at hand instead of the gold standard, we can ensure that medical findings better serve the global community.

\section*{Acknowledgements}
This work was supported, in part, by the Institute of Education Sciences, U.S. Department of Education, through Grant \textit{R305D220032} to Harvard University and Bentley University. The authors gratefully acknowledge the Laboratory for Computational Physiology for sharing data and providing comments and feedback in the early stages of this project. We also thank workshop participants at IIT Demokritos, the APHP, the Japanese Society of Intensive Care Medicine (especially Satoru Hashimoto and Hide Shigemitsu), the National University of Singapore Saw Swee Hock School of Public Health, the Harvard Data Science Initiative Conference, the Asian Political Methodology Conference, and the 2023 American Causal Inference Conference, especially Rohit Bhattacharya and Alex D'Amour.

 \section*{Data Availability}

 The data used in this study are from the MIMIC-III database, which is freely-accessible for authorized users. Researchers interested in obtaining access must submit a formal request at: \url{https://mimic.physionet.org/gettingstarted/access/}. The data used for the analyses presented throughout this paper are, however, available upon request to individuals who present proof of MIMIC-III permissions.

\section*{Code Availability}
All software tools R functions used to implement the methods discussed in this paper are publicly available at \url{https://github.com/reaganmozer/textmatch}. Code and replication materials for the EHR study are available upon request for individuals who present proof of MIMIC-III permissions.

	\clearpage

	\bibliographystyle{apalike}
	\bibliography{refs.bib}
	
\section*{Author contributions}	
	RM and AK contributed to the methodological development, data analysis, interpretation of results, and writing of the manuscript. LC contributed to the preparation of the EHR study data and interpretation of results. LM contributed to the development of methods, interpretation of results, and writing of the manuscript. All authors read and approved the final manuscript.
\end{document}